\documentclass{INTERSPEECH2023}

\usepackage{amsmath,graphicx,listings,xcolor}
\usepackage{multirow}
\usepackage{hyperref}

\interspeechcameraready

\title{Turbo your multi-modal classification with contrastive learning}
\name{Zhiyu Zhang$^{\star}$, Da Liu$^{\star}$, Shengqiang Liu, Anna Wang, Jie Gao, Yali Li$^{\dagger}$ \thanks{${\star}$ Equal contributions}\thanks{${\dagger}$ Corresponding author}}
\address{NIO}
\email{yali.li@nio.com}

\begin{document}

\maketitle
 
\begin{abstract}
Contrastive learning has become one of the most impressive approaches for multi-modal representation learning. However, previous multi-modal works mainly focused on cross-modal understanding, ignoring in-modal contrastive learning, which limits the representation of each modality. In this paper, we propose a novel contrastive learning strategy, called $Turbo$, to promote multi-modal understanding by joint in-modal and cross-modal contrastive learning. Specifically, multi-modal data pairs are sent through the forward pass twice with different hidden dropout masks to get two different representations for each modality. With these representations, we obtain multiple in-modal and cross-modal contrastive objectives for training. Finally, we combine the self-supervised Turbo with the supervised multi-modal classification and demonstrate its effectiveness on two audio-text classification tasks, where the state-of-the-art performance is achieved on a speech emotion recognition benchmark dataset.

\end{abstract}

\section{Introduction}

Recently, as an effective self-supervised strategy of learning representations, contrastive learning has been successfully applied in multi-modal tasks. By learning the alignment between different modalities, contrastive learning projects different modal representations into a joint semantic space. For example, CLIP \cite{radford2021learning} uses large-scale image text pairs to learn uni-modal representations and bring the representations into the joint multi-modal space through contrastive learning. Similarly, CLAP \cite{elizalde2022clap} learns to associate language and audio modalities by contrastive learning. They have achieved outstanding results comparable to supervised learning. 

The general steps of previous multi-modal contrastive learning methods can be summarized as follows. First, the two modalities are modeled by their respective encoders to obtain the corresponding representation. Then the respective representations from the same multi-modal pair constitute the positive samples, and the rest in the mini-batch constitute negative samples. Finally, contrastive learning is used to learn representations by pulling positive samples closer together and pushing negative samples farther away. However, this approach only focuses on the contrast of cross-modal representations, ignoring in-modal contrastive learning. The success of uni-modal contrastive learning methods, such as SimCLR \cite{chen2020simple} and SimCSE \cite{gao2021simcse}, has proven that in-modal contrastive learning can effectively enhance representation learning. Accordingly, we propose simultaneously incorporating in-modal and cross-modal contrastive learning to promote multi-modal understanding. On the other hand, previous multi-modal contrastive learning approaches are usually used in pre-training, which requires enormous amounts of pair-wise multi-modal data. However, collecting and cleaning pair-wise data is difficult, especially in specific domains. Therefore, we adopt a simple approach that directly combines the supervised multi-modal classification task with contrastive learning.

In this paper, we propose a novel self-supervised contrastive learning method called $Turbo$ and apply it to multi-modal classification. Motivated by the idea of R-Drop \cite{wu2021r} and SimCSE \cite{gao2021simcse}, we use $dropout$  \cite{srivastava2014dropout} twice to get multiple representations for two modalities and utilize these representations to compare and learn in-modal and cross-modal information. Dropout is usually a way to regularize a network, but here we are using its randomness to get different representations for data augmentation. Specifically, in the training step, we let each multi-modal data pair go through the forward pass twice and get two different representations for each modality by randomly dropping out some hidden units. Then we carry out in-modal contrastive learning for the two representations from the same modality and cross-modal contrastive learning for the representation of different modalities. The loss of the Turbo we defined is the sum of several in-modal and cross-modal contrastive losses. Finally, we train Turbo as an auxiliary task to improve the performance of the multi-modal classification tasks. Comparing to previous multi-modal methods, Turbo first propose to incorporate in-modal contrastive learning and obtains multiple in-modal and cross-modal contrastive objectives from the same input pair, enhancing the generalization of representation learning. 

By combining the Turbo training method, we have achieved significant improvements in audio-text classification tasks. Compared to the baseline, we observed accuracy improvements of 5.59\% for the speech emotion recognition task and 3.83\% for the device-directed speech detection task. At the same time, we outperform the previous systems on the emotion recognition benchmark dataset IEMOCAP \cite{busso2008iemocap} and achieve state-of-the-art performance. Further analysis of experimental results shows that the Turbo method can improve the $alignment$ and $uniformity$ \cite{wang2020understanding} of modal representations, thus improving the performance of the system.

\section{Method}

\begin{figure*}[htb]
\includegraphics[scale=0.7]{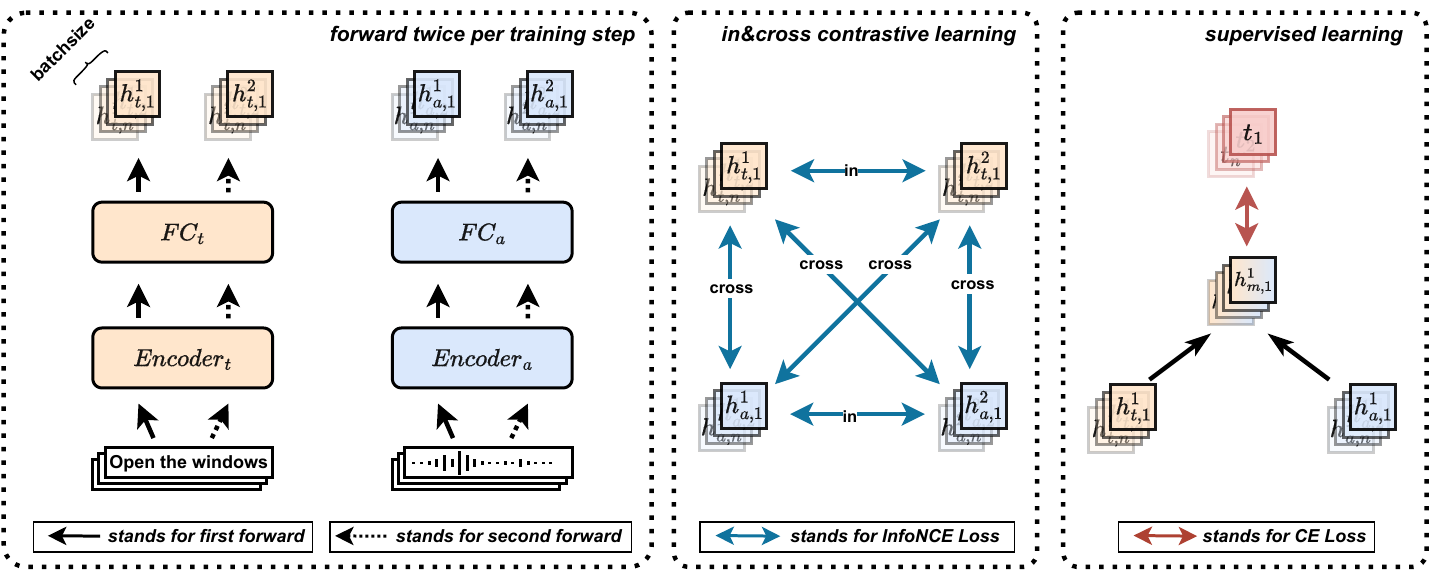}
\centering
\caption{Overview of our proposed classification framework with Turbo}
\label{fig:Turbo}
\end{figure*}

The overall framework of our model with the Turbo method is shown in Fig.\ref{fig:Turbo}. We take acoustics and language modalities as an example to introduce the proposed multi-modal framework. Audio-text pairs are firstly sent into an audio encoder, e.g., wav2vec2 \cite{baevski2020wav2vec} or Whisper \cite{radford2022robust}, and a text encoder, e.g., BERT \cite{devlin2018bert} or GPT \cite{radford2019language}, respectively. Then we construct multiple in-modal and cross-modal contrastive objectives for audio-text pairs in a mini-batch using the Turbo method. Concurrently, representations of audio-text pairs will be concatenated and sent to a linear classifier. Finally, we jointly optimize the classifier's supervised loss and the Turbo's self-supervised loss. We show numpy-like pseudocode for the Turbo in Listing 1.

\subsection{Encoder with Dropout Mask}
For each input audio-text pair $\{{x}_{\text{a}}, {x}_{\text{t}}\}$, we first feed the audio and text to the encoder respectively to get the utterance-level embedding. Let $\mathbf{e}_{\text{a}}\in {{\mathbb{R}}^{d1}}$ denote the audio embedding, and $\mathbf{e}_{\text{t}}\in {{\mathbb{R}}^{d2}}$ denote the text embedding after the encoder. Each audio-text pair can be denoted as $\{\mathbf{e}_{\text{a}}, \mathbf{e}_{\text{t}}\}_{i}$ in a mini-batch, where ${i}\in [0,N]$  and $N$ is the batch size. Then we send audio and text embeddings, $\mathbf{e}_{\text{a}}$ and $\mathbf{e}_{\text{t}}$, into a joint multi-modal space of dimension $d$ by using a trainable fully-connected layer:
\begin{equation}\label{eq1}
{{\mathbf{h}}_{\text{a}}}={FC}_{\text{a}}(\mathbf{e}_{\text{a}}); {{\mathbf{h}}_{\text{t}}}={FC}_{\text{t}}(\mathbf{e}_{\text{t}})
\end{equation}

\noindent
where ${{\mathbf{h}}_{\text{a}}}\in{{\mathbb{R}}^{d}}$, ${{\mathbf{h}}_{\text{t}}}\in{{\mathbb{R}}^{d}}$, ${FC}_{\text{a}}$ and ${FC}_{\text{t}}$ are the fully-connected layers for audio and text respectively. 
After the above process, we can obtain $\{\mathbf{h}_{\text{a}}, \mathbf{h}_{\text{t}}\}_{i}$ for the audio-text pair $\{{x}_{\text{a}}, {x}_{\text{t}}\}_{i}$ through a training forward pass.

Inspired by SimCSE \cite{gao2021simcse} and R-Drop \cite{wu2021r}, \emph{dropout} can be used in data augmentation and regularizing the output predictions. We apply it to multi-modal contrastive learning. Concretely, we place dropout masks on the encoders and fully-connected layers. Dropout will randomly drop part of units in each neural network layer.  We feed the same input audio-text pair $\{{x}_{\text{a}}, {x}_{\text{t}}\}_{i}$ to the forward pass of the network twice. Because of different hidden dropout masks, we can get two representation pairs: $\{\mathbf{h}_{\text{a}}^{1}, \mathbf{h}_{\text{t}}^{1}\}_{i}$ and $\{\mathbf{h}_{\text{a}}^{2}, \mathbf{h}_{\text{t}}^{2}\}_{i}$.

\subsection{In-modal and Cross-modal Contrastive Learning}
Previous works, such as CLIP and CLAP, usually focus on cross-modal contrastive learning and ignore in-modal information, weakening the ability to extract information from a single modality. Consequently, we add the extra in-modal contrastive learning in the training step. Taking the audio modality as an example, $\mathbf{h}_{\text{a}}^{1}$ and $\mathbf{h}_{\text{a}}^{2}$ are similar audio representations obtained from the two forward passes. Then we take them as “positive pairs,” and other audio representations in the same mini-batch as “negatives” during contrastive learning. The specific training objective for in-modal contrastive learning can be defined as follows:
\begin{equation}\label{eq2}
\ell_{in}^{a}=-\log \sum_{i=1}^N  \frac{\operatorname{exp}\left({\operatorname{sim}\left(\mathbf{h}_{a,i}^{1}, \mathbf{h}_{a,i}^{2}\right) / \tau}\right)}{\sum_{j=1}^N \operatorname{exp}\left({\operatorname{sim}\left(\mathbf{h}_{a,i}^{1}, \mathbf{h}_{a,j}^{2}\right) / \tau}\right)}
\end{equation}

\noindent
where $\ell_{in}^{a}$ is in-modal InfoNCE loss \cite{oord2018representation} of the audio modality, $N$ is the batch size, $\tau$ is a temperature hyper-parameter, ${\operatorname{sim}(.)}$ denotes the cosine similarity calculation and $\mathbf{h}_{a,i}^{1}, \mathbf{h}_{a,i}^{2}$ represent the audio representations for first and second forward passes respectively. For the text modality, we apply the same process to get $\ell_{in}^{t}$.

Since there are two audio-text representation pairs $\{\mathbf{h}_{\text{a}}^{1}, \mathbf{h}_{\text{t}}^{1}\}_{i}$ and $\{\mathbf{h}_{\text{a}}^{2}, \mathbf{h}_{\text{t}}^{2}\}_{i}$ for the same pair $\{{x}_{\text{a}}, {x}_{\text{t}}\}_{i}$, we can build four different cross-modal contrastive objectives, i.e. $\{\mathbf{h}_{\text{a}}^{1}, \mathbf{h}_{\text{t}}^{1}\}_{i}$, $\{\mathbf{h}_{\text{a}}^{1}, \mathbf{h}_{\text{t}}^{2}\}_{i}$, $\{\mathbf{h}_{\text{a}}^{2}, \mathbf{h}_{\text{t}}^{1}\}_{i}$, $\{\mathbf{h}_{\text{a}}^{2}, \mathbf{h}_{\text{t}}^{2}\}_{i}$. Then we can respectively calculate the corresponding cross-modal InfoNCE Loss: $\ell_{cross}^{1,1}$, $\ell_{cross}^{1,2}$, $\ell_{cross}^{2,1}$ and $\ell_{cross}^{2,2}$. Taking $\ell_{cross}^{1,1}$ as an example, it means the contrastive loss between $\mathbf{h}_{\text{a}}^{1}$ and $\mathbf{h}_{\text{t}}^{1}$ in the mini-batch, which can be defined as:
\begin{equation}\label{eq3}
\ell_{cross}^{1,1}=-\log \sum_{i=1}^N  \frac{\operatorname{exp}\left({\operatorname{sim}\left(\mathbf{h}_{a,i}^{1}, \mathbf{h}_{t,i}^{1}\right) / \tau}\right)}{\sum_{j=1}^N \operatorname{exp}\left({\operatorname{sim}\left(\mathbf{h}_{a,i}^{1}, \mathbf{h}_{t,j}^{1}\right) / \tau}\right)}
\end{equation}

Then, incorporating in-modal and cross-modal contrastive learning, the loss of Turbo is formulated as:
\begin{equation}\label{eq4}
\ell_{Turbo}= \ell_{in}^{a}+\ell_{in}^{t}+\ell_{cross}^{1,1}+\ell_{cross}^{1,2}+ \ell_{cross}^{2,1}+\ell_{cross}^{2,2}
\end{equation}

\subsection{Supervised Classification with Turbo}
For each labeled audio-text pair, we obtain two representation pairs  $\{\mathbf{h}_{\text{a}}^{1}, \mathbf{h}_{\text{t}}^{1}\}_{i}$ and $\{\mathbf{h}_{\text{a}}^{2}, \mathbf{h}_{\text{t}}^{2}\}_{i}$ after two forward passes. Each pair can be chosen to be trained for supervised classification. Taking $\{\mathbf{h}_{\text{a}}^{1}, \mathbf{h}_{\text{t}}^{1}\}_{i}$ as an example, we concatenate it and send it through a simple linear classifier. After that, the predicted probability distribution $\mathbf{\hat{y}}$ for the target class is as follows:
\begin{equation}\label{eq5}
\mathbf{\hat{y}}=\text{softmax}[\,\mathbf{W}\cdot (\mathbf{h}_{\text{a}}^{1}\oplus \mathbf{h}_{\text{t}}^{1})+\mathbf{b}\,]
\end{equation}

\noindent
where $\mathbf{W}$ and $\mathbf{b}$ are trainable parameters, and $\oplus$ denotes the concatenation operation. Multi-class cross-entropy (CE) is chosen as the loss function of this classifier: $\ell_{ce}$.

Finally, we treat Turbo as an auxiliary task for classification. The overall objective will be:
\begin{equation}\label{eq6}
\ell_{total}= \lambda \ell_{ce}+(1-\lambda)\ell_{Turbo}
\end{equation}
where $\lambda$ is a balancing hyperparameter.

\section{Experiments}
\subsection{Datasets}

To evaluate the proposed multi-modal model, we conduct experiments on two audio-text classification tasks, i.e., speech emotion recognition (SER) and device-directed speech detection (DSD).

\noindent
\textbf{Speech Emotion Recognition:} The purpose of the SER task is to identify whether the audio recording belongs to one of the categories, such as happy, sad, angry, or neutral. We use the Interactive Emotional Dyadic Motion Capture (IEMOCAP) \cite{busso2008iemocap}, an open multi-modal emotion recognition benchmark dataset. Following previous works  \cite{krishna2020multimodal} \cite{liu2020group}, the dataset contains 5,531 utterances in total (1,636 happy, 1,084 sad, 1,103 angry and 1,708 neutral). In the training process, we perform 10-fold cross-validation where each 8, 1, and 1 folds are used for the train, validation, and test sets, respectively. The performance of this task is measured in widely used evaluation metrics: weighted accuracy (WA) which is the overall classification accuracy and unweighted accuracy (UA) which is the average accuracy over the  emotion categories. 

\noindent
\textbf{Device-directed Speech Detection:} The DSD task aims to distinguish voice queries intended for a virtual voice assistant device from background speech \cite{mallidi2018device} \cite{norouzian2019exploring}. Since this is no publicly available data for this task, we evaluate it on the real-life in-house dataset, named REJ, which consists of audio utterances with ground truth annotations of device-directed or non-device-directed. The dataset contains roughly 50K utterances and is split into train, validation, and test partitions with 40K, 5K, and 5K utterances, respectively. The class split is roughly 1:1 to balance the two classes in each partition. We utilize the equal error rate (EER) and accuracy (ACC) to measure the classification.

\lstset{  
 commentstyle=\color[RGB]{0,96,96},     
 basicstyle=\footnotesize\ttfamily,
 language=python,                       
}

\begin{lstlisting}[frame=tb,caption= Numpy-like pseudocode for the core of an implementation of Turbo., captionpos=b,belowcaptionskip=2cm, label=zebra]
# audio_encoder - Wav2vec2 or Whisper
# text_encoder - Bert or GPT

# extract feature representations
A_1st = audio_encoder(audio_input)
T_1st = text_encoder(text_input)

# second forward calculation
A_2nd = audio_encoder(audio_input)
T_2nd = text_encoder(text_input)

# in-modal contrastive learning
loss += InfoNce(np.dot(A_1st, A_2nd))
loss += InfoNce(np.dot(T_1st, T_2nd))

# cross-modal contrastive learning
loss += InfoNce(np.dot(A_1st, T_1st))
loss += InfoNce(np.dot(A_1st, T_2nd))
loss += InfoNce(np.dot(A_2nd, T_1st))
loss += InfoNce(np.dot(A_2nd, T_2nd))

\end{lstlisting}

\begin{table}[htbp]
\centering
\begin{tabular}{l|cccc}
\hline
\toprule
 \multirow{2}{*}{Method} & \multicolumn{2}{c}{IEMOCAP} & \multicolumn{2}{c}{REJ}\\
\cmidrule(lr){2-3}  \cmidrule(lr){4-5}
 & WA & UA & ACC & EER \\
 \midrule 
$Vanilla$ & 0.7076 & 0.7177 & 0.8944 & 0.0938 \\ 
$+CL_{cross}$  & 0.7238 & 0.7265 & 0.9152 & 0.0786\\ 
$+Turbo$  & \textbf{0.7635} &  \textbf{0.7709} &  \textbf{0.9327} & \textbf{0.0676}\\ 
\bottomrule
\end{tabular}
\caption{Comparison of baselines and our method}
\label{tab1}
\end{table}

\begin{table}[htbp]
\centering
\begin{tabular}{l|c|c|c}
\hline
\toprule
 Method & Year & WA & UA\\
\midrule
Xu et al. \cite{xu2019learning}& 2019  & 0.725 & 0.709 \\ 
Liu et al. \cite{liu2020group}& 2020  & 0.724 & 0.701\\ 
Krishna et al. \cite{krishna2020multimodal}& 2020  & - & 0.728\\ 
Li et al. \cite{li2020learning}& 2020  & 0.727 & 0.735\\ 
Makiuchi et al. \cite{makiuchi2021multimodal}& 2021  & 0.735 & 0.730\\ 
Chen et al. \cite{chen2022key}& 2022 & 0.743 &0.753\\
Ours& 2023  & \textbf{0.764} &  \textbf{0.771}\\ 
\bottomrule
\end{tabular}
\caption{Comparison with previous methods on the IEMOCAP}
\label{tab2}
\end{table}

\subsection{Experimental Setup}

We use wav2vec 2.0-base and BERT-base models from HuggingFace repositories$\footnote{\url{https://huggingface.co}}$ as the audio and text encoders, which both have 768-dimensional embeddings. The optimizer for the model is Adam, with a learning rate of 1e-5. The training batch size is 64, and we set the early stopping setting as ten epochs. Dropout is applied with a probability of 0.2 after every feed-forward layer except the output layer. The hyperparameter $\lambda$ is set to 0.5 according to experimental results. All experiments are conducted on a single Nvidia RTX 3090 GPU. 

For comparison, we build two multi-modal models as our baselines. The first one, the vanilla model, directly concatenates the output of audio and text encoders for classification. It forwards only once and without any contrastive learning loss. The second one is based on the vanilla model with a simple cross-modal contrastive learning loss, similar to CLAP\cite{elizalde2022clap}. 

\subsection{Results}
Table \ref{tab1} shows the results of the mentioned approaches for the IEMOCAP and REJ datasets. $+CL_{cross}$ denotes the second baseline with cross-modal contrastive loss. Overall, the proposed framework with the Turbo method substantially outperforms the vanilla model, a strong baseline based on the pre-trained wav2vec and BERT in both tasks. For the IEMOCAP dataset, the proposed Turbo achieves improvements of 5.59\% and 5.32\% on UA and WA. Equally, Turbo improves the accuracy by 3.83\% and reduces EER by 2.62\% for the REJ dataset, confirming the proposed method's effectiveness and universality. In addition, comparing the second baseline with the vanilla model, we can find that cross-modal contrastive learning improves performance for multi-modal classification. This is because the contrastive loss will guide the classification loss during training and affect the representations to improve audio-text alignment. Compared with the two baselines, Turbo yields stable improvements no matter what dataset, which shows the superiority of our special in-modal and cross-modal contrastive learning method to understand multi-modal information.

The comparison results of our method with some previous multi-modal methods on the IEMOCAP are listed in Table \ref{tab2}. These methods have learned modal information by directly calculating the correlation between different modalities based on the attention mechanism, which can generate extra noise information and damages the performance. They concentrated on the complex strategy of fusing multiple modalities, without considering the distribution of the representations within each modality. In contrast, we use Turbo contrastive learning method to guide the representations of different modalities to aligned feature space. The experimental results indicate that the proposed method can improve the ability of the fusion of modalities without any complex fusion methods (we only use the method of concatenating features directly) and achieve state-of-the-art performance. 

\subsection{Analysis}

To better understand the effectiveness of Turbo, we use $alignment$ and $uniformity$ \cite{wang2020understanding} to measure the quality of learned representations. The pseudocode for these two metric calculations is shown in listing 2, which is slightly different from the uni-modal computing method in \cite{wang2020understanding}. $Alignment$ means the alignment (closeness) between corresponding cross-modal audio-text pairs, and $uniformity$ means the uniformity (coverage) of the uni-modal representation space on the unit hypersphere. In general, models with better alignment and uniformity can achieve better performance. 

\begin{figure}[htb]
\includegraphics[scale=0.6]{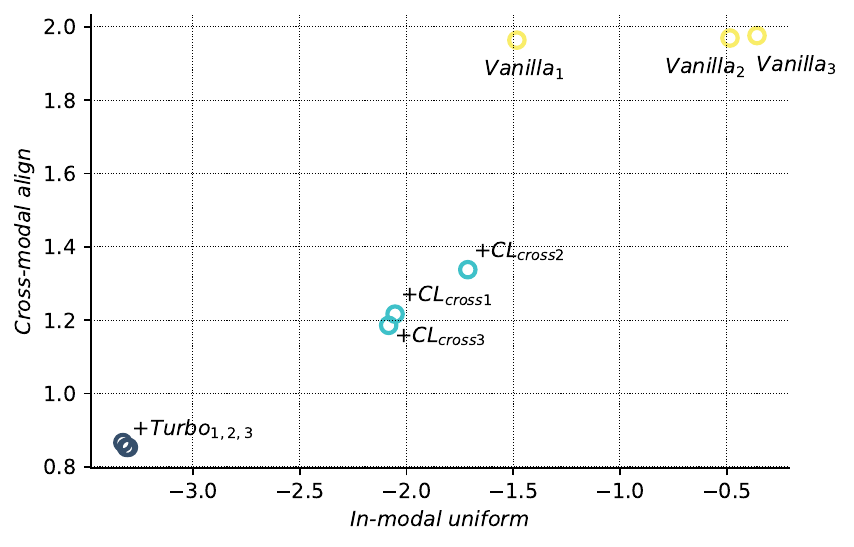}
\centering
\caption{The $align$-$uniform$ plot of models}
\label{fig:alignuniform}
\end{figure}

\begin{figure}[htp]
\includegraphics[scale=0.22]{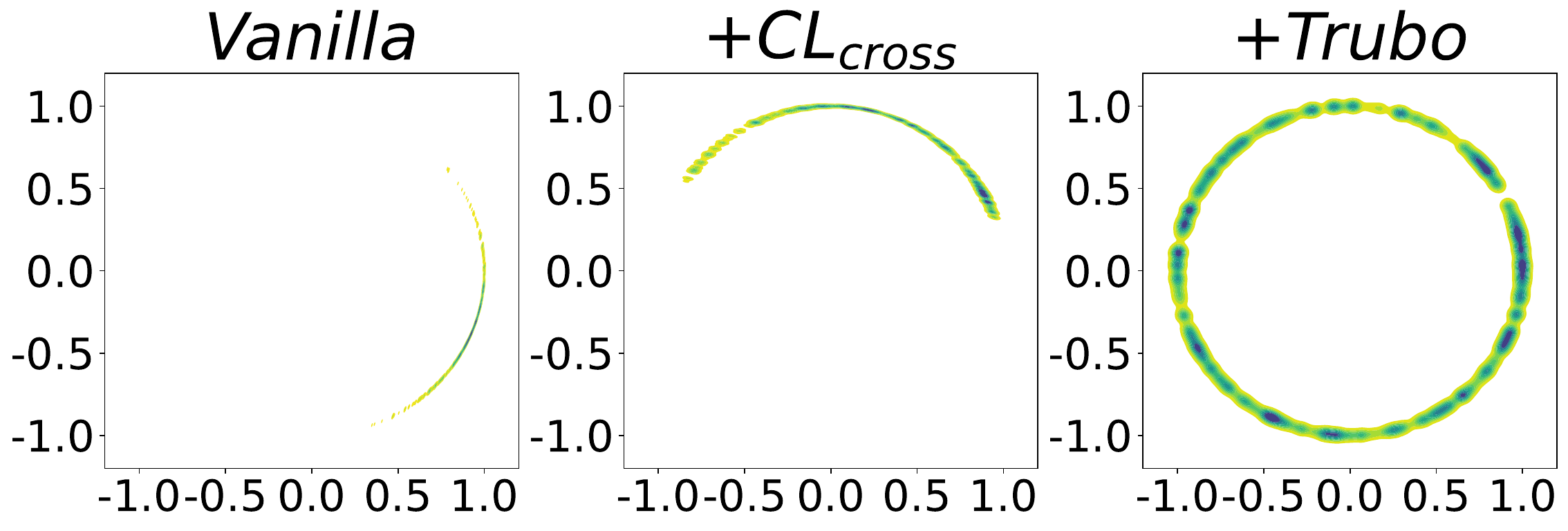}
\centering
\caption{Feature distributions with Gaussian kernel density estimation (KDE) in $\mathbb{R}^{2}$.}
\label{fig:circle}
\end{figure}

We show the alignment and uniformity plot of three models evaluated on the IEMOCAP in Fig.\ref{fig:alignuniform}. Lower numbers of alignment and uniformity are better. For each method, we train three times and visualize the embeddings in the figure. There are several observations. Compared with the vanilla model, the "$+CL_{cross}$" model dramatically improves the alignment because of the cross-modal contrastive learning. Turbo can further improve the alignment since it has multiple cross-modal contrastive objectives for the same audio-text pair. In terms of uniformity, the Turbo method is substantially superior to the "$+CL_{cross}$" and vanilla models. That is because Turbo performs additional in-modal contrastive learning. This phenomenon is also reflected in Fig.\ref{fig:circle}, in which we visualize the distributions of audio representation space for three models. Results indicate that the Turbo method achieves the most uniform distribution, which again proves the superiority of our method.

\begin{lstlisting}[frame=tb,caption= Pytorch-like pseudocode for calculating alignment and uniformity., captionpos=b,belowcaptionskip=2cm, label=zebra]
# A: the distribution of audio representation
# T: the distribution of text representation

# the calculation of alignment
def lalign(A, T, alpha=2):
    return (A-T).norm(dim=1).pow(alpha).mean()

# the calculation of uniformity
def lunif(A, t=2):
    sq_pdist = torch.pdist(A, p=2).pow(2)
    return sq_pdist.mul(-t).exp().mean().log()
    
\end{lstlisting}

\section{Conclusions}
In this work, we propose Turbo, a contrastive learning strategy for multi-modal classification tasks. Unlike previous works that only focus on cross-modal contrastive learning, Turbo performs in-modal contrastive learning simultaneously. By forwarding twice, Turbo obtains multiple in-modal and cross-modal contrastive objectives for the same input pair, which enhances the generalization of representation learning. We demonstrate the effectiveness of our method on two audio-text classification tasks. Compared with the vanilla method, we achieved a 5.59\% improvement in weighted accuracy on the SER task and a 3.83\% improvement in weighted accuracy on the DSD task. Further analysis indicates that Turbo can effectively enhance the uniformity and alignment of multi-modal representations. 

In the future, we will explore extending our proposed Turbo to audio-text pre-trained learning and improve performance for the downstream tasks. We also intent to expand our proposed multi-modal learning method to include the visual domain and integrate three modalities for classification.

\section{Acknowledgements}

\ifinterspeechfinal
     The INTERSPEECH 2023 organisers
\else
     The authors
\fi
would like to thank ISCA and the organising committees of past INTERSPEECH conferences for their help and for kindly providing the previous version of this template.

\newpage

\bibliographystyle{IEEEtran}
\bibliography{main}

\end{document}